\pdfoutput=1

\documentclass[11pt]{article}

\PassOptionsToPackage{hyphens}{url}
\usepackage[hyperref]{acl}

\usepackage{times}
\usepackage{latexsym}

\usepackage{booktabs}
\usepackage{graphicx}
\usepackage{tablefootnote}

\usepackage[T1]{fontenc}

\usepackage[utf8]{inputenc}

\usepackage{microtype}

\urldef{\donnaStricklandUrl}\url{https://wikimediafoundation.org/news/2018/10/04/donna-strickland-wikipedia/#:~:text=Donna\%20Strickland\%20is\%20an\%20optical,of\%20a\%20Sloan\%20Research\%20Fellowship.}

\title{Generating Full Length Wikipedia Biographies\\The Impact of Gender Bias on the Retrieval-Based \\ Generation of Women Biographies}

\author{Angela Fan \\
  FAIR / LORIA \\
  Université de Lorraine\\
  \texttt{angelafan@fb.com} \\\And
  Claire Gardent \\
  CNRS/LORIA \\
  Nancy, France\\
  \texttt{claire.gardent@loria.fr} \\}

\begin{document}
\maketitle
\begin{abstract}
Generating factual, long-form text such as Wikipedia articles raises three key challenges: how to gather relevant evidence, how to structure information into well-formed text, and how to ensure that the generated text is factually correct. 
We address these by developing a model for English text that uses a retrieval mechanism to identify relevant supporting information on the web 
and a cache-based pre-trained encoder-decoder to generate long-form biographies section by section, including citation information. To assess the impact of available web evidence on the output text, we compare the performance of our approach when 
generating biographies about women (for which less information is available on the web) vs. biographies generally.  
To this end, we curate a dataset of 1,500 biographies about women.
We analyze our generated text to understand how differences in available web evidence
data affect generation. We evaluate the factuality, fluency, and quality of the generated texts using automatic metrics and human evaluation. We hope that these techniques can be used as a starting point for human writers, to aid in reducing the complexity inherent in the creation of long-form, factual text. 
\end{abstract}

\section{Introduction}

Wikipedia has become one of the major sources of dissemination of knowledge across the globe. 
However, the knowledge contained in Wikipedia is not neutral --- it is biased in various  ways~\cite{hinnosaar2019gender,schmahl2020wikipedia}.
Many studies, including those from the Wikimedia Foundation itself, have emphasized that biographies in particular are overwhelmingly written about men. 
This leads to many subtle yet far-reaching effects, from students not writing their first book reports on a woman to bias in models trained on Wikipedia, as Wikipedia has long been used as a source of data. 
Many existing efforts, such as the Wikipedia Women in Red project, focus on encouraging article creation to mitigate this gender gap.
However, Wikipedia articles remain painstakingly written and edited primarily by a network of human contributors. 
Despite advances in text generation and modeling architectures that retrieve information, the automatic creation of Wikipedia articles is incredibly challenging~\cite{liu2018generating}.
Even the functionality of tools that aid human editors are limited. 

In this work, we strive to create a system that could write an entire Wikipedia article in English, focusing on the biography domain. 
We confront several major challenges. First, this is fundamentally a long-form generation task. Improvements driven by pretraining~\cite{radford2019language,lewis2019bart} have improved generation fluency at the level of multiple sentences. However, Wikipedia biographies contain multiple paragraphs in a structured form with headings, as well as citations to indicate where the information originated from. Second, the task confronts obstacles around the factuality~\cite{elazar2021measuring} of generated content, as articles must be factually accurate. Third, Wikipedia articles are written using reference material, often found on the web~\cite{piktus2021web}. Thus, models need to find and ingest web searches as a pre-requisite to writing accurate biographies. 

We develop a method for English Wikipedia that starts with the subject and occupation of the biography, then leverages web search to find relevant evidence. Given search results, we employ a retrieval-augmented generation architecture~\cite{lewis2020retrieval,guu2020realm} based on large-scale pretraining to identify relevant information and write the biography. We generate section by section, using a caching mechanism similar to Transformer-XL~\cite{dai2019transformer} to reference previous sections and achieve greater document-level context. Finally, after each section, we append a citation based on which web searches were retrieved. 

We quantify the quality of generation using several automatic metrics such as ROUGE-L~\cite{lin2004rouge}, entailment, and named entity coverage. 
Further, we study the strong dependency of our method on accurate retrieval, and design a specific evaluation dataset that highlights this challenge. The dataset consists of 1,527 Wikipedia biographies about women, where information on the internet is not as easily retrieved. We use this dataset to analyze the gap between model quality when retrieval is challenging (our novel evaluation dataset with biographies about women) and model quality when retrieval is more accurate (a random set of evaluation biographies). 
Finally, we conduct a large-scale human evaluation to measure the factuality and coverage of our generated biographies. We hope that our techniques can eventually be used as a starting point for human Wikipedia writers, for biographies and beyond.

\section{Related Work}

\subsection{Generation of Wikipedia Articles}

A large body of work in generation utilizes Wikipedia, often for data-to-text tasks that use Wikidata or DBpedia RDF triples~\cite{gardent-etal-2017-webnlg,castro-ferreira-etal-2020-2020,kaffee2018mind,vougiouklis2018neural,sha2018order,puduppully2019data,chen2020kgpt,wang2020towards,agarwal2020large,parikh2020totto}, as well as graphs~\cite{jin2020genwiki} as input. 
Some have focused on long text, such as writing summaries~\cite{chen2020generating} or sections of articles~\cite{kaffee2020using}, expanding stubs~\cite{banerjee2015wikikreator}, and writing full articles~\cite{liu2018generating}. Some of these  works utilize structure to learn templates~\cite{sauper2009automatically}, Markov logic networks~\cite{liu2010biosnowball}, or word graphs~\cite{banerjee2015wikikreator}, but we anticipate that pretraining and large neural network based techniques will vastly improve upon this quality.

Closest to our work, \citet{liu2018generating} use web evidence to write full length articles, but do not focus on biographies and use extractive summarisation techniques rather than a retrieval mechanism to identify relevant information. Further, their work generates the entire Wikipedia article at once, whereas we demonstrate that breaking down the article to generate section by section is more effective. We also include a mechanism for the model to generate citations, which was not included in existing work. Thus, our model can produce a full-form Wikipedia article that would look like what a human editor wrote.

Finally, our work (i) leverages recent advances in large-scale pretraining, which improves generation fluency and (ii) investigates the impact of available web evidence on the generated texts.

Other work has focused on automatic creation of biographies, such as generation from infoboxes~\cite{lebret2016neural} or Wikidata~\cite{chisholm2017learning}, as well as extracting biographical sentences~\cite{biadsy2008unsupervised}. The majority of existing research focused on short biographies.

\subsection{Retrieval in Generative Models}

Retrieval mechanisms have been used to support a variety of tasks, including dialogue~\cite{moghe2018towards,dinan2018wizard,shuster2021retrieval}, fact verification~\cite{thorne2018fever}, and sentence generation~\cite{guu2018generating}. Most notably, retrieval has been heavily used in question answering~\cite{chen2017reading,kwiatkowski2019natural,seo2019real,karpukhin-etal-2020-dense}. Recent innovations in incorporating retrieval mechanisms have increased the quality and scale of retrieval-augmented generative methods~\cite{guu2020realm,lewis2020retrieval,izacard2020leveraging}. 

\subsection{Bias in Wikipedia Biographies}

Gender bias on Wikipedia is a well-known problem~\cite{hinnosaar2019gender,dinan2020multi,schmahl2020wikipedia}, particularly in the case of biographies~\cite{graells2015first,stratigakos2016unforgetting,luo2018ladies,schmahl2020wikipedia}. This bias is compounded by geographical location, as information about certain areas of the world is far more prevalent~\cite{kaffee2018learning,beytia2020positioning}. This bias exists not only in what articles are written, but also in articles targeted for deletion --- articles about certain marginalized groups are removed at higher rates~\cite{worku2020exploring}. 
Wikipedia reflects biases present in society~\cite{de2019bias,young2020gender,schmahl2020wikipedia}, though numerous initiatives exist to de-bias Wikipedia. These range from training programs~\cite{iglesias2020preparing} to projects such as Women in Red\footnote{\url{https://en.wikipedia.org/wiki/Wikipedia:WikiProject_Women_in_Red}} and WikiProject Women\footnote{\url{https://en.wikipedia.org/wiki/Wikipedia:WikiProject_Women}}. The success of these initiatives has been studied~\cite{langrock2020gender} and found to be effective, but not at addressing the systemic challenges that create bias in the first place. 

In the natural language processing community, work has focused on combating gender bias in co-reference resolution~\cite{zhao2018gender}, dialogue~\cite{dinan2019queens,lee2019exploring,liu2020mitigating}, detection of abusive language~\cite{park2018reducing}, machine translation~\cite{stanovsky2019evaluating}, and word embeddings~\cite{gonen2019lipstick}. These works present a variety of strategies, including data augmentation, additional data collection efforts, modified generation, and fair evaluation~\cite{yeo2020defining}. A comprehensive survey can be found in~\citet{blodgett2020language}. However, most of these efforts are focused on specific tasks or models --- our work uniquely targets generation of full Wikipedia biographies to combat gender bias present on Wikipedia.

\begin{figure*}[t]
    \centering
    \includegraphics[width=\linewidth]{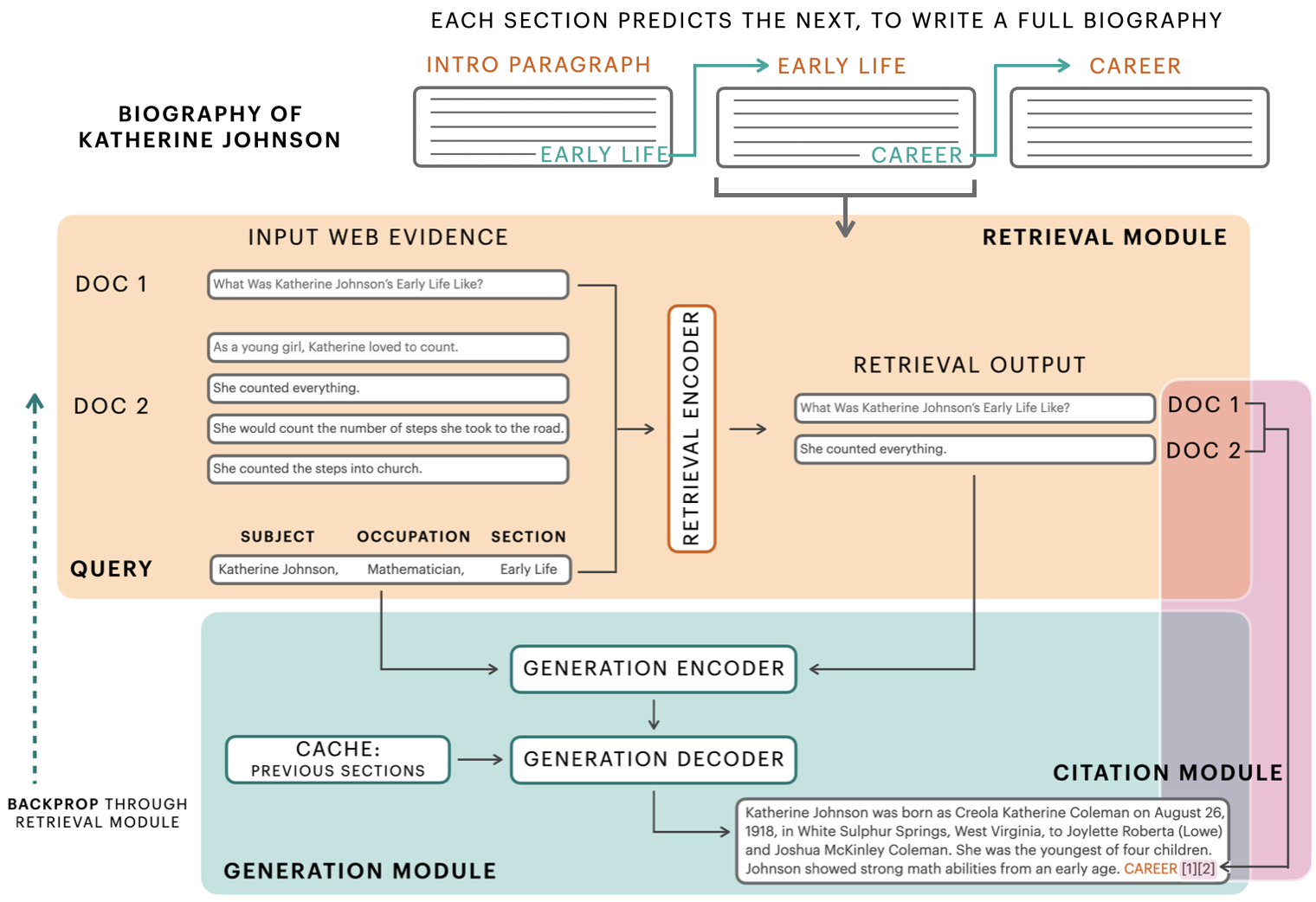}
    \caption{\textbf{Model Architecture.} Our method writes a Wikipedia article section by section, with each section predicting the next in sequence. 
    To write one section, the model starts with a \textit{retrieval module} that uses a query consisting of the subject name, occupation, and section heading to identify the most relevant information from the web. The query and retrieval output passes to the \textit{generation module}, which generates the desired section while using a cache to reference previously written sections. Finally, to complete the full Wikipedia article, the \textit{citation module} appends citations based on the retrieved content. The entire system is learned end-to-end, with backpropagation from the generation module through the retrieval module.}
    \label{fig:architecture}
\end{figure*}

\section{Task}
\label{sec:task}

Given a person's name, one or more occupation(s), and CommonCrawl as a source of evidence, the task is to generate a Wikipedia biography and to associate each generated section with adequate bibliographic references. 
We model this task by generating a biography section by section using  section headers as additional information. A special section header called \textit{toplevel} is used as the start of the article. The subsequent headers are automatically generated at the end of each section as input for the next. Thus for each section, the input includes a name, one or more occupations, a section header, and CommonCrawl as a retrieval corpus. 

\section{Method}
\label{sec:method}

Wikipedia biographies begin with an introductory paragraph followed by various subsections\footnote{Many biographies contain infoboxes, which we do not generate.}. 
To account for this structure and generate long-form text based on retrieved web evidence, our system, illustrated in Figure~\ref{fig:architecture}, generates a biography section by section. Based on the subject, their occupation(s), and the section heading, the model first identifies  a subset of relevant evidence from a set of web search results found using that triplet (\textit{retrieval module}). It then conditions upon that evidence to generate the section, using a Sequence-to-Sequence model (\textit{generation module}) which can access previous sections using a caching mechanism. Finally, the model indicates which evidence documents it used and outputs those as citations, mimicking a standard Wikipedia article (\textit{citation module}). We focus on generation in English.

\subsection{Retrieval Module}

Given  a query $Q$ and a set of web documents  $D$ retrieved from the web based on this query, the task of the retrieval module is to retrieve the subset of $D$ that is most relevant given $Q$. The challenge is sifting through the large quantity of potentially useful information. 

\paragraph{Query.} The query $Q$ consists of three parts: \textbf{(1)} the name of the person for which the biography is generated, \textbf{(2)} their , possibly multiple, occupation(s), and \textbf{(3)} a section heading. Including the occupation narrows the realm of potential relevant content, especially as proper names are often ambiguous (e.g. \textit{Jane Wang}). Similarly, the section header allows the model to retrieve different information for each section (e.g. \textit{Personal Life} compared to \textit{Career}).

\paragraph{Documents.} The query $Q$ is put through a search engine to retrieve web hits, which form the set of documents $D$ that are candidates for retrieval. The web results are represented only as text, and all non-text information is discarded.

\paragraph{Retrieval.} 
To retrieve the relevant subset of $D$, each sentence in $D$ is encoded with RoBERTa base trained with LayerDrop~\cite{fan2019reducing,liu2019roberta,devlin2018bert}. The concatenation of the subject's name, occupation(s), and section header is also encoded.
We then calculate the dot product to identify which encoded document sentences are most relevant given the currently encoded query $Q$, following the strategy used in other retrieval works \cite{karpukhin-etal-2020-dense}.
The representation of the top $k$ most relevant sentences are then passed onwards through the model.
Note that compared to some other retrieval-augmented generation~\cite{lewis2020retrieval}, the RoBERTa encoder is not fixed, so the retrieval module learns based on the performance of the generation module.
This is possible because our retrieval is far smaller scale, we limit the search to approximately 40 sentences (1,000 words) that could be used to generate each section.

\subsection{Generation Module}

To generate the sections 
we use a Transformer-based Sequence-to-Sequence model initialized with BART-Large~\cite{lewis2019bart}. 
The input to BART is the concatenation of the subject's name, occupation(s), the section header and the retrieved evidence. Note that the maximum number of input tokens for BART is 1024 words, which is why we cap the retrieval at 1000 words, as described in the previous section. 
The decoder conditions on the input information to generate the section.

One challenge with this is that the sections would be generated completely independently, which 
might result in redundancy  between generated sections.
Thus, we equip the Sequence-to-Sequence model with a mechanism to refer to previous sections using the cache mechanism from Transformer-XL~\cite{dai2019transformer}. This mechanism caches the previous section's hidden states at every layer, using it as memory to generate the current section.

\subsection{Citation Module}

Recent work has focused on models that not only perform a task, but also produce an explanation~\cite{deyoung2019eraser}. Much of this work has focused on question answering~\cite{latcinnik2020explaining,lamm2020qed,lakhotia2020fid,gonzalez2020human} and generating explanations in natural language~\cite{camburu2019make,narang2020wt5,kumar2020nile,hase2020leakage}. A similar requirement exists on Wikipedia --- not only to collate the information into an article, but to provide the \textit{original references} for users to verify. Thus, to complete the generation of a full Wikipedia biography, we cite the information used, as in any real article. On Wikipedia itself, each sentence could contain citations. We simplify this, citing at the end of each section. To do this, we track the original document the retrieved evidence originates from, and reference that document at the end of the generated section.

\subsection{Bringing it All Together}

To write a full biography, models must generate the introductory paragraph followed by each section. For a new article, the introductory paragraph is given as a section heading called \textit{toplevel}. For each subsequent section, we follow the process outlined above to retrieve evidence, then write a section, then add citations. At the end of each section, the model generates the \textit{section heading} of the next section. This allows the model to generate an entire article section by section.

\section{Creating an Evaluation Dataset}

A possible failure point for our method is the retrieval step as good biography generation requires access to sufficient relevant information. 
To study the impact of accurate retrieval on generation quality, we design a specific evaluation dataset that pushes this problem to the forefront.
Specifically, we create a novel evaluation dataset which consists exclusively of biographies about women. 

Ongoing efforts to write biographies about women in the Wikipedia editor community, such as the Women in Red project, have identified \textit{insufficient online evidence} as a major challenge for writing Wikipedia biographies about women. 
To study the importance of retrieval on model quality, we therefore 
create an evaluation dataset where the target Wikipedia articles are women bios. 
We collate candidate biographies, retrieve information about their occupation, and gather web sources using web search. The resulting dataset, summarized in Table~\ref{table:eval_dataset_stats}, consists of 1,527 biographies, each linked to a set of retrieved web articles.

\paragraph{Identifying Biographical Subjects.}

We first source various notable women on Wikipedia using internet lists (e.g. \textit{Famous Women you should know}) and existing efforts by collective groups of Wikipedia editors, such as the Women in Red project. Several recent efforts focus on Women in Science~\footnote{\url{https://towardsdatascience.com/who-is-wikipedia-famous-within-natural-language-processing-fa0c8e91bdf6?gi=b910dd838c47},\url{https://www.newscientist.com/article/mg24532680-800-jess-wades-one-woman-mission-to-diversify-wikipedias-science-stories/}}, and so we specifically include scientists as a category. Overall, we collate almost two thousand candidate Wikipedia women biographies. We then narrow down by selecting articles that have previously \textit{Featured Article} or \textit{Good} quality. The final evaluation dataset contains 1,527 biographies in four groups: Women, Women in Science, Women in Asia, and Women in Africa (see Table~\ref{table:eval_dataset_stats}).

\begin{table*}[]
\centering \small
\begin{tabular}{ll}
\toprule
\bf Most Common Section Headings & Career, Personal Life, Early Life, Biography, History \\
\bf Most Common Occupations &  Writer, Politician, University Teacher, Physician, Researcher\\
\bottomrule
\end{tabular}
\caption{\label{table:example_occupations} \textbf{Example Section Headings and Occupations in Wikipedia Biographies.} }
\end{table*}

\begin{table}[]
\centering \small
\begin{tabular}{lc}
\toprule
\bf WikiSum Evaluation Dataset & \\ 
\midrule \midrule 
Average Number of Sections & 7.2 \\ 
Average Length of a Section & 151.0 \\ 
Average Length of Total Article & 892.3 \\ 
\midrule
Avg overlap of Web Hits and Biography & 39.8\% \\ 
\toprule
\bf Our Evaluation Dataset & \\ 
\midrule 
\midrule 
Average Number of Sections & 5.8 \\ 
Average Length of a Section & 132.3 \\ 
Average Length of Total Article & 765.9 \\ 
\midrule 
Avg Number of Web Hits (max 20) & 18.1 \\ 
Avg overlap of Web Hits and Biography & 24.9\% \\ 
\midrule
Biographies about Women & 419 \\ 
Biographies about Women in Science & 808 \\ 
Biographies about Women in Asia & 164 \\ 
Biographies about Women in Africa & 136 \\ 
Total Biographies & 1,527 \\
\bottomrule
\end{tabular}
\caption{\label{table:eval_dataset_stats} \textbf{Breakdown and Statistics of Biographies} of a random sample of Wikipedia biographies compared to our created evaluation dataset.}
\end{table}

\paragraph{Biography Text and Occupation.}

After finalizing candidate Wikipedia biographies, we use the MediaWiki  API\footnote{\url{https://www.mediawiki.org/wiki/API}} to query the text of the article. We use the Wikidata API\footnote{\url{https://query.wikidata.org/}} to retrieved the individuals, possibly multiple, occupations 
(e.g. \textit{Rachel Carson} is an author and an environmental activist). As seen in Table~\ref{table:eval_dataset_stats}, on average, articles have around 6 sections with 130 words each. The most common occupations include writers, teachers, and doctors (see Table~\ref{table:example_occupations}), though the entire dataset contains almost 500 different occupations, with people having on average 2 occupations (see Table~\ref{table:eval_dataset_stats}). 

\paragraph{Retrieving Web Evidence.}

Next, we identify web sources with reference evidence for each biography. We follow the construction of similar datasets, such as WikiSum~\cite{liu2018generating} and ELI5~\cite{fan2019eli5}, which searches  through CommonCrawl. We query CommonCrawl based on the subject's name and occupation(s) and return the top 20 search results~\cite{shuster2022language,komeili2021internet}. We reject all CommonCrawl links from Wikipedia, to prevent querying the Wikipedia articles in our dataset. Statistics are presented in Table~\ref{table:eval_dataset_stats}. Out of a maximum of 20 possible hits, on average each biography returns around 18.

\section{Experimental Details}

We describe our training data, baselines, and automatic and human evaluation metrics.

\paragraph{Training Data.} 

We utilize the WikiSum~\cite{liu2018generating} dataset of Wikipedia articles paired with web references. We filter to biographies using a combination of querying for occupations in Wikidata and using Named Entity Recognition\footnote{\url{https://spacy.io/usage/linguistic-features/}} to recognize names. We query each article title in the WikiSum dataset to attempt to find an occupation and see the title is recognized as a named entity, to identify the bibliographical subset of WikiSum. This produces 677,085 biographies, each associated with a set of web articles. 

\paragraph{Evaluation Data.} 

We utilize the WikiSum~\cite{liu2018generating} dataset, filtered to biographies, for evaluation. Similar to the training dataset, we query to identify occupational information.
To study the impact of retrieval and available evidence on model quality, we also evaluate on our constructed evaluation dataset about women (which has substantially less web-based evidence). As shown in Table~\ref{table:eval_dataset_stats}, these two datasets differ in the length and quality of both the Wikipedia articles and the web-based evidence. 

\paragraph{Baseline.} 

We compare our method described in Section~\ref{sec:method} to a pretraining and finetuning generation baseline. We use the BART model~\cite{lewis2019bart} and finetune on the Biography subset of the WikiSum data. 
Note that BART has a token limit of 1024, thus the entirety of the web retrieval is not available to this model. We take the web search hits ordered by the search engine, and provide the first 1000 available tokens.
To compare this baseline with our method equitably, the baseline is also trained to generate section by section. However, it does not use the retrieval module (all evidence is given), the caching mechanism, or the citation module (as described in Section~\ref{sec:method}), meaning citations are not added to the generated text. Additional training details are in the Appendix.

\paragraph{Generation.}

We generate from all models with beam search, setting the beam size to 5. We allow the model to generate an output of any length, with no restrictions. For human evaluations, we set the minimum and maximum length such that it matches the length of the gold target to minimize the effect of length on human interpretations.

\paragraph{Automatic Evaluation.}

We evaluate the quality of generated biographies with three automatic metrics. First, we measure the \textbf{ROUGE-L} between the generated text and the 
Wikipedia reference text to assess the similarity. ROUGE-L is commonly used in multi-sentence summarization and is a measure of longest common substring overlap.

Next, we use \textbf{Natural Language Entailment} as a high level proxy for quantifying a form of factuality: if two sentences entail each other in both directions, then they are semantically equivalent. 
We use a model pretrained and finetuned on MNLI, open sourced by~\citet{liu2019roberta}. To evaluate entailment, we split the generated biography and reference biography into sentences, then for each sentence in the generated biography we calculate if it is semantically equivalent to a sentence in the reference. We then compute the percentage of generated sentences that are semantically equivalent to at least one sentence in the reference biography, where entailment is evaluated bidirectionally.

Finally, we assess the \textbf{Coverage} of information in the generated biography, constraining this to analyzing mentions of named entities. We report the percentage of named entities detected in the reference which are also detected in the generated text. We extract entities with \textsc{blink}, a BERT-based entity linking system~\cite{wu2019scalable}.  

\paragraph{Human Evaluation}

Long-form text generation is very difficult to assess automatically~\cite{thomson2020gold,howcroft2020twenty}, particularly for factuality~\cite{goodrich2019assessing,maynez2020faithfulness,peshterliev2021conversational} and hallucination~\cite{zhou2020detecting,duvsek2020evaluating}. We conduct a detailed, large-scale human evaluation with the goal to assess \textbf{Coverage} (How much of the information in the reference section is in the generated section?) and \textbf{Factuality} (How much of the generated section is in the reference and, for the information added in the generated text, how much of that information is verifiable based on the 
web evidence?). 

To reduce the challenge of evaluation, the text is compared section by section, and the generated text is the same length as the reference by constraining the max length of beam search (to remove length as an evaluation artifact).
First, each sentence of the generated section is shown next to the full reference section and the entire document cited in the generated section (recall our generated biographies cite the retrieved evidence). Evaluators are asked to decide \textbf{(1)} if the information in the generated sentence is present in the reference section (ground truth) and \textbf{(2)} if the information in the generated sentence is present in the cited document (web evidence). This question assesses if the information from the generated section is factual with respect to either the reference Wikipedia text or the retrieved web documents.
Then, the evaluation is flipped to assess coverage with respect to the Wikipedia reference. Each sentence of the reference is shown next to the generated section, and evaluators are asked to decide \textbf{(3)} if the information in the reference sentence is present in the generated section. In total, human annotators evaluated 100 sections with length between 200 to 500 words. Each section is reviewed by one annotator. Additional details are in the Appendix. 

\section{Results and Discussion}

We describe our main results and analyze the importance of retrieval on model quality. An example generation is shown in Figure~\ref{fig:example_wikipedia_generation}.

\begin{table*}[]
\centering \small
\begin{tabular}{lccc}
\toprule
\bf Model & \bf ROUGE-L & \bf Entailment & \bf Named Entity Coverage \\ 
\midrule 
BART Pretraining + Finetuning & 17.4 & 15.8 & 21.9 \\ 
+ Retrieval Module & 18.8 & 17.2 & 23.1 \\ 
+ Caching Mechanism & 19.3 & 17.9 & 23.4 \\ 
\bottomrule
\end{tabular}
\caption{\label{table:full_results} \textbf{Full Results on Biography Generation.} We compare the BART baseline with our method across different automatic metrics to assess fluency, factuality, and coverage. Results are shown on the test set.}
\end{table*}

\definecolor{mypink}{RGB}{199, 82, 154}
\definecolor{mygreen}{RGB}{54, 150, 137}
\definecolor{myorange}{RGB}{189, 96, 46}
\newcommand{\pink}[1]{\textcolor{mypink}{#1}}
\newcommand{\green}[1]{\textcolor{mygreen}{#1}}
\newcommand{\orange}[1]{\textcolor{myorange}{\underline{#1}}}

\begin{figure*}[t]
  \small
  \rule{\linewidth}{1pt}
    \green{hyman is best known for her work on the classification of invertebrates}. \green{she was the author of a six-volume set of reference books titled the invertebrate} \pink{treatise}, \pink{which was published by mcgraw-hill} \orange{in the united states and in germany}. \green{she also wrote a series of laboratory manuals for the teaching of zoology classes nationwide}. \pink{hyman's work has had a lasting influence on scientific thinking} \orange{about a number of animal groups, and the only works that can be compared with hers are of} \green{composite authorship}.
  \rule{\linewidth}{1pt}
  \caption{\textbf{Example Generation} of the \textit{Work} section for a biography about Libbie Hyman, a zoologist. Green indicates text in the reference article, Pink indicates text in the web evidence, and Orange (underlined) indicates hallucination. See the biography on Wikipedia: \url{https://en.wikipedia.org/wiki/Libbie_Hyman}.}
  \label{fig:example_wikipedia_generation}
\end{figure*} 

\begin{table}[]
\centering \small
\begin{tabular}{lc}
\toprule
\bf Model & \bf ROUGE-L \\ 
\midrule 
\multicolumn{2}{l}{\bf Retrieval with Different Queries} \\ 
with Subject Name Only & 19.6 \\ 
with Name and Occupation & 19.8 \\ 
with Name, Occupation, Section Heading & 21.4 \\ 
\midrule 
\multicolumn{2}{l}{\bf Writing Articles in Sections} \\ 
Entire Article & 14.4 \\ 
Section by Section & 15.9 \\ 
\bottomrule
\end{tabular}
\caption{\label{table:ablation} \textbf{Ablations} of types of Queries for the Retrieval Module and generation section by section. Results are shown on the dev set.}
\end{table}

\subsection{Quality of Generated Biographies}

\paragraph{Automatic Evaluation.} We examine the model's overall performance. Results are summarized in Table~\ref{table:full_results}. Compared to the pretraining+finetuning baseline, adding the retrieval module statistically significantly\footnote{We use the confidence interval reported in the ROUGE package.} increases results by 1.4 ROUGE-L. Adding a caching mechanism improves further by 0.5 ROUGE-L. This trend is reflected across the entailment and entity coverage metrics, indicating that retrieving the most relevant information to write a biography is critical. 

Next, we examine the impact of our modeling choices using ablation (Table~\ref{table:ablation}). Compared to previous work on WikiSum~\cite{liu2018generating,fan2019using}, we add an end-to-end retrieval mechanism based on RAG that substantially improves results. Further, instead of retrieving solely based on the subject name, as was previously done~\cite{liu2018generating}, we retrieve on a detailed query (the name, occupation, and section heading). Table~\ref{table:ablation} indicates that this enriched query improves the retrieval quality by almost 2 ROUGE-L. We conjecture it helps improve disambiguation and retrieve evidence that is relevant to the desired entity rather than to one of its homonyms.

We also generate the biographical articles section by section, rather than an entire article at once. This allows the retrieval mechanism to be focused on the section information. As shown in Table~\ref{table:ablation}, this also has a positive effect of +1.5 ROUGE-L. 

\begin{figure}[t]
    \centering
    \includegraphics[width=\linewidth]{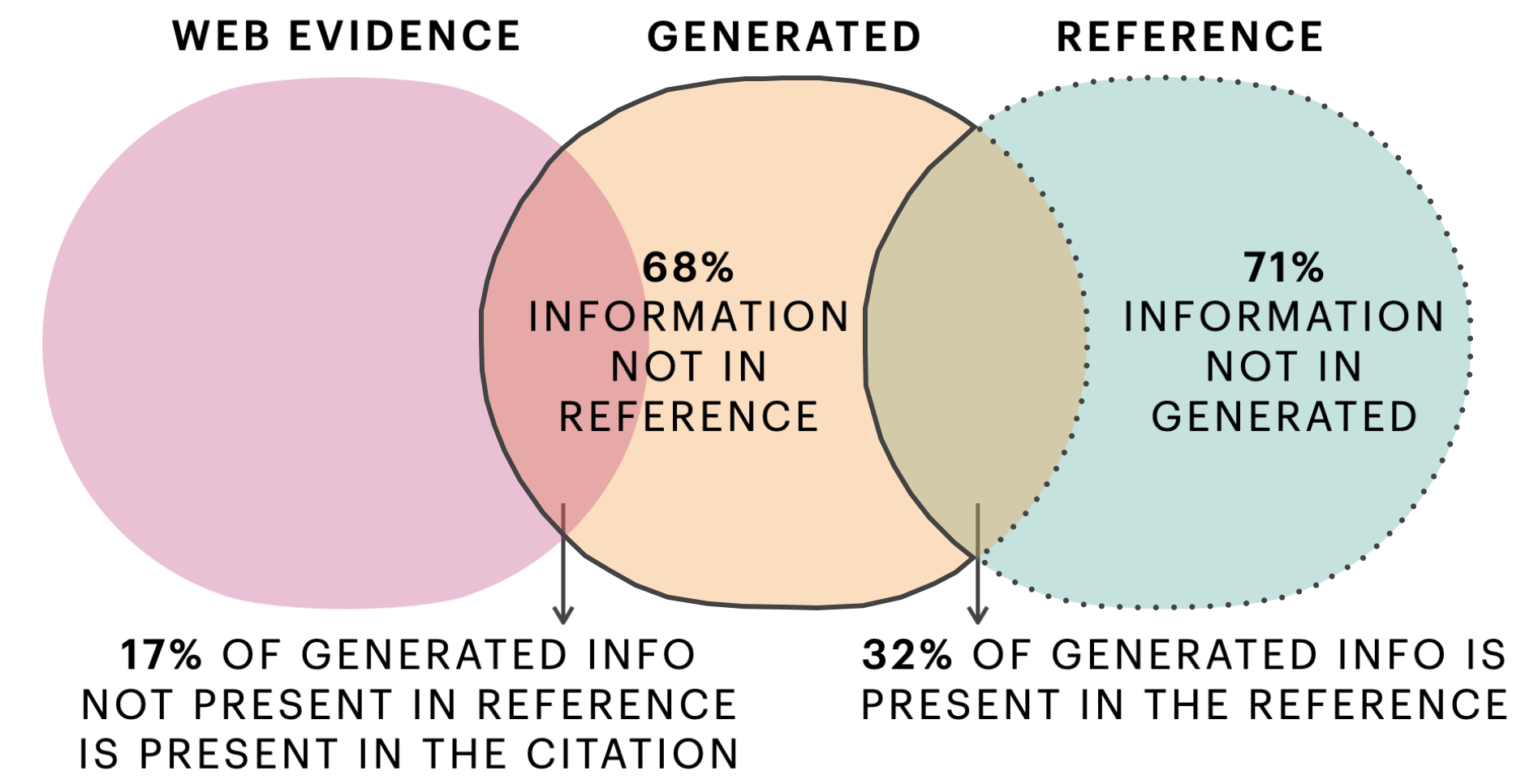}
    \caption{\textbf{Human Evaluation.} We compare the coverage of content between generated and reference biographies, as well as the factuality of generated content.}
    \label{fig:human_eval}
\end{figure}

\begin{table*}[]
\centering \small
\begin{tabular}{lccccc}
\toprule
\bf Model & \bf WikiSum Test & \bf Women & \bf Scientists & \bf Women in Asia & \bf Women in Africa \\ 
\midrule 
BART Pretraining & 19.0 & 17.4 & 18.2 & 16.7 & 16.4  \\ 
+ Retrieval & 21.4 & 18.8 & 19.3 & 17.9 & 17.1 \\ 
+ Caching & 21.8 & 19.3 & 19.7 & 18.4 & 17.3 \\ 
\bottomrule
\end{tabular}
\caption{\label{table:subset_results} \textbf{ROUGE-L Performance broken down by sub-categories.} We compare the BART baseline with our method across different subsets of women, as well as the biography subset of WikiSum Test.}
\end{table*}

\begin{table}[]
\centering \small
\begin{tabular}{lccc}
\toprule
\bf Model & \bf WikiSum & \bf Women  & \bf Women  \\ 
 & \bf Test & \bf  Asia & \bf  Africa \\ 
\midrule 
Our Method & 19.0 & 16.7 & 16.4  \\ 
+ finetune on Women & 18.9 & 17.3 & 16.8  \\ 
\bottomrule
\end{tabular}
\caption{\label{table:finetuning_results} \textbf{Improved Performance when Finetuning} on biographical articles with less web evidence. We finetune on biographies about women that do not include this subset of women in Asia and Africa.}
\end{table}

\paragraph{Human Evaluation.} Next, we examine quality with human evaluation, as shown in Figure~\ref{fig:human_eval}. Models generating nonfactual or hallucinated content is an ongoing area of study~\cite{tian2019sticking,nie2019simple,liu2021token}. Our goal is to understand how much information in the generated text is present in the reference text or the web evidence, as a proxy for factuality and coverage. 
Overall, 68\% of the information in generated sections is not present in the reference text. Conversely, 71\% of information in the reference text is not in the generated text. This  indicates that the generated text has far from perfect coverage. However, 
we found that 17\% of the added information can be validated by examining the web evidence, which shows that some information added by the generative model is valid biographical information. 

We examine why there is low information overlap between the generated and reference text. First, information in the reference biography may not be available on the web\footnote{Note that search hits from the Wikipedia domain are removed from web search results.} or may not be retrieved. In a manually examined subset of 250 sentences taken from reference biographies, we found that about 50\% of the information was not contained in the web evidence. The other 50\% was partially present in the web evidence but were not retrieved by the model. Second, annotators must compare sentences, but sentences contain partial information. For example, if \textit{Person is was born in Chicago in 1968} was in the generated text and \textit{Person was born in Chicago} was in the reference text, this would count as the generation having information not in the reference. 
Annotators were very precise in sticking to the requested standard that the \textit{entire sentence} should be factual to count as fully factual, which is reflected by annotators marking \textit{partial} factuality as not factual. Our stringent standard for factuality produces a clearer understanding of hallucinations at the sentence-level.

In summary, our investigation suggests two explanations for the low coverage reported by human annotators: lack of information in the web evidence and difficulty assessing whether two sentences contain the same core knowledge.

\subsection{Performance with Unreliable Retrieval}
One major challenge of accurate Wikipedia article generation is when information is not available on the web or not easily retrieved.
For example, information could simply not exist on the internet. Writing a Wikipedia biography about any randomly chosen person on the street would likely manifest this scenario. 
Other situations could include having a large number of search results returned but difficulty identifying which are relevant,  having too few search results to write a good biographic article, or even having only noise returned in the search results.
We discuss these challenges and possible mitigations in this section. 

\paragraph{The Evidence Gap.} 

We compare the results on our evaluation set about women with those on the WikiSum test set. 
Compared to WikiSum, the unigram overlap of the web hits with the biographical article is substantially lower for our evaluation dataset (see Table~\ref{table:eval_dataset_stats}).
As shown in Table~\ref{table:subset_results}, across the board, the quality of generated biographies is higher for the WikiSum Test set. This is especially prominent for Women in Asia and Africa, which are more than 2.5 ROUGE-L worse on average. 

\paragraph{Reducing the Dependency on Retrieval.}

One challenge is that there is a disconnect between the training dataset, where retrieval information is readily available, and the women-focused evaluation dataset, where retrieval information is noisy or missing. 
We investigate the potential of a straightforward strategy to mitigate differences in training data: that of training on biographical articles with less reliable web evidence. We mimic this by finetuning our model on a subset of our evaluation dataset, and then testing on Women in Asia and Africa, the two categories that perform most poorly. As shown in Table~\ref{table:finetuning_results}, finetuning statistically significantly improves performance, though the improvement is not large (+0.5 ROUGE-L). 
Another phenomenon that arises with noisy web evidence is that retrieving more is not necessarily better. Perhaps only one website has really relevant information.
In the retrieval module, all available web documents are encoded at the sentence level, and the model can select sentences across all documents.
We next explore an approach where the model first scores documents, then selects sentences from the most relevant document. 
We found this had very similar performance, and thus conclude that the challenge of identifying relevant documents and then sentences is probably similar in difficulty to identifying relevant sentences directly.

\section{Conclusion}

We developed a novel retrieval and cache-augmented generative model to generate long-form biographies based on evidence from the web. Experimental evidence reveals that an enriched query including occupations, caching, and backpropagation through the retrieval module contributes to improved performance. We investigate the dependency on high-quality web evidence, which manifests strongly in our constructed evaluation dataset of biographies about women. We discuss this challenge and possible mitigations.

\section{Acknowledgments}

We thank the anonymous reviewers for their feedback. We gratefully acknowledge the support of the French National Research Agency and of Facebook AI Research Paris (for Claire Gardent; award ANR-20-CHIA-0003, XNLG "Multi-lingual, Multi-Source Text Generation").

We thank Adina Williams, Emily Dinan, Ledell Wu, and Aleksandra Piktus for thoughtful discussions and feedback on this entire effort, as well as previous collaborations that influenced this work. We thank Sebastian Riedel, Douwe Kiela, Mona Diab, and Michael White for their suggestions to improve this work. We thank Mojtaba Komeili for developing the web query service we used to create the evaluation dataset.

Finally, we thank all of the editors of Wikipiedia, particularly those in the Women in Red Project, for their hard work and dedication to creating, moderating, editing, and all that is necessary to keep Wikipedia running. We encourage readers to donate to Wikipedia to support this public project.

\section{Ethical Considerations}
\label{sect:ethics} 

In this section, we discuss several known limitations and ethical considerations of our work. We do not recommend any kind of text generation technology to be deployed on Wikipedia given this is an active area of research. 

\subsection{Dependency on Evidence from the Web reflects Bias on the Internet}

Biographies, whether written as books or available online, reflect societal bias. While many Wikipedia editors rely on web-based references to create their articles, and we follow the same strategy in this work, relying on the web is flawed. The prominent reason is that the internet is full of bias in it of itself. For example, Donna Strickland, who received a Nobel Prize, did not have a Wikipedia article\footnote{\donnaStricklandUrl}
as there was not sufficient content about her on the web as a basis for her article. Thus, it is important to recognize that the availability  of references is problematic, affecting the downstream ability to write accurate, comprehensive biographies. Further, information on the web can be contradictory, information can be affected by the passage of time, and not information on the web is necessarily factually correct. Our proposed modeling mechanism does not have a way to explicitly recognize or correct for these challenges, which also plagues text generation generally. 

\subsection{Focus on English Limits Inclusivity from Other Languages}

Our work focuses on text generation in English only, which limits inclusivity purely on the basis of language. This is challenging as the content of the internet and Wikipedia itself is different in various languages. For example, articles about people from Germany may be more likely to be located on the German version of Wikipedia. Another factor is that the content of the references may be written in another language, and then used by a bilingual individual to write an article in English about that subject. This is often the case for many biographical subjects who may be more well known in a non-English speaking area. 

\subsection{Evaluation focuses on Women Only, Not Other Groups}

There are a very large number of marginalized groups in the world and numerous important intersectional aspects to consider. When discussing identity, a wide variety of factors and personal views influence individuals when thinking about how they describe themselves. Our evaluation dataset focuses on women alone, which leaves out many groups, including non-binary people. Further, Wikipedia may not reflect the up-to-date information --- names and gender are both mutable, for example --- and Wikipedia articles do not ask each subject to self-report their gender. Finally, we note that by grouping people into hard categories, there can potentially be harm --- such as limiting people from opportunities because of their gender or race. However, we strongly believe that it is important to recognize bias in its various forms as it exists, particularly in popular, default online sources of information such as Wikipedia.

\subsection{Bias in Style, Word Choice, and Tone}

In this work, we focus on bias manifesting as unequal prevalence and length of biographical content on Wikipedia, focusing specifically on different intersectional groups of women. However, bias manifests in a number of other ways. Studies have indicated that the words used in biographies about women compared to biographies about men~\cite{dinan2019queens} also differs, and is reflective of gendered terminology. For example, many articles about women are actually written with a lot of information about men, such as their husband's careers, and articles about actresses describe more often their physical appearance. This is also a manifestation of bias, and we do not present any focused modeling techniques to address this type of bias explicitly. 

\subsection{Biographies as Records}

In the modern internet, a large number of events are recorded for the public record. These include events that people may personally prefer to forget, often termed \textit{right to be forgotten}\footnote{\url{https://en.wikipedia.org/wiki/Right_to_be_forgotten}}. Automatically generating biographies about individuals may collate such information in an easily accessible public place, which can conflict with this personal right. This has a complex but important interaction with marginalized groups. For example, many celebrities who are women, transgender, or a part of another marginalized group are far more likely to have news articles written about intimate personal details such as plastic surgeries. Thus, it is important to consider the interaction of biographical data with individual privacy. This is a larger challenge of biographical information generally.

\bibliography{anthology,custom}

\begin{thebibliography}{87}
\expandafter\ifx\csname natexlab\endcsname\relax\def\natexlab#1{#1}\fi

\bibitem[{Agarwal et~al.(2020)Agarwal, Ge, Shakeri, and
  Al-Rfou}]{agarwal2020large}
Oshin Agarwal, Heming Ge, Siamak Shakeri, and Rami Al-Rfou. 2020.
\newblock Large scale knowledge graph based synthetic corpus generation for
  knowledge-enhanced language model pre-training.
\newblock \emph{arXiv preprint arXiv:2010.12688}.

\bibitem[{Banerjee and Mitra(2015)}]{banerjee2015wikikreator}
Siddhartha Banerjee and Prasenjit Mitra. 2015.
\newblock Wikikreator: Improving wikipedia stubs automatically.
\newblock In \emph{Proceedings of the 53rd Annual Meeting of the Association
  for Computational Linguistics and the 7th International Joint Conference on
  Natural Language Processing (Volume 1: Long Papers)}, pages 867--877.

\bibitem[{Beyt{\'\i}a(2020)}]{beytia2020positioning}
Pablo Beyt{\'\i}a. 2020.
\newblock The positioning matters: Estimating geographical bias in the
  multilingual record of biographies on wikipedia.
\newblock In \emph{Companion Proceedings of the Web Conference 2020}, pages
  806--810.

\bibitem[{Biadsy et~al.(2008)Biadsy, Hirschberg, and
  Filatova}]{biadsy2008unsupervised}
Fadi Biadsy, Julia Hirschberg, and Elena Filatova. 2008.
\newblock An unsupervised approach to biography production using wikipedia.
\newblock In \emph{Proceedings of ACL-08: HLT}, pages 807--815.

\bibitem[{Blodgett et~al.(2020)Blodgett, Barocas, Daum{\'e}~III, and
  Wallach}]{blodgett2020language}
Su~Lin Blodgett, Solon Barocas, Hal Daum{\'e}~III, and Hanna Wallach. 2020.
\newblock Language (technology) is power: A critical survey of" bias" in nlp.
\newblock \emph{arXiv preprint arXiv:2005.14050}.

\bibitem[{Camburu et~al.(2019)Camburu, Shillingford, Minervini, Lukasiewicz,
  and Blunsom}]{camburu2019make}
Oana-Maria Camburu, Brendan Shillingford, Pasquale Minervini, Thomas
  Lukasiewicz, and Phil Blunsom. 2019.
\newblock Make up your mind! adversarial generation of inconsistent natural
  language explanations.
\newblock \emph{arXiv preprint arXiv:1910.03065}.

\bibitem[{Castro~Ferreira et~al.(2020)Castro~Ferreira, Gardent, Ilinykh,
  van~der Lee, Mille, Moussallem, and
  Shimorina}]{castro-ferreira-etal-2020-2020}
Thiago Castro~Ferreira, Claire Gardent, Nikolai Ilinykh, Chris van~der Lee,
  Simon Mille, Diego Moussallem, and Anastasia Shimorina. 2020.
\newblock \href {https://aclanthology.org/2020.webnlg-1.7} {The 2020 bilingual,
  bi-directional {W}eb{NLG}+ shared task: Overview and evaluation results
  ({W}eb{NLG}+ 2020)}.
\newblock In \emph{Proceedings of the 3rd International Workshop on Natural
  Language Generation from the Semantic Web (WebNLG+)}, pages 55--76, Dublin,
  Ireland (Virtual). Association for Computational Linguistics.

\bibitem[{Chen et~al.(2017)Chen, Fisch, Weston, and Bordes}]{chen2017reading}
Danqi Chen, Adam Fisch, Jason Weston, and Antoine Bordes. 2017.
\newblock Reading wikipedia to answer open-domain questions.
\newblock \emph{arXiv preprint arXiv:1704.00051}.

\bibitem[{Chen et~al.(2020{\natexlab{a}})Chen, Wiseman, and
  Gimpel}]{chen2020generating}
Mingda Chen, Sam Wiseman, and Kevin Gimpel. 2020{\natexlab{a}}.
\newblock Generating wikipedia article sections from diverse data sources.
\newblock \emph{arXiv preprint arXiv:2012.14919}.

\bibitem[{Chen et~al.(2020{\natexlab{b}})Chen, Su, Yan, and
  Wang}]{chen2020kgpt}
Wenhu Chen, Yu~Su, Xifeng Yan, and William~Yang Wang. 2020{\natexlab{b}}.
\newblock Kgpt: Knowledge-grounded pre-training for data-to-text generation.
\newblock \emph{arXiv preprint arXiv:2010.02307}.

\bibitem[{Chisholm et~al.(2017)Chisholm, Radford, and
  Hachey}]{chisholm2017learning}
Andrew Chisholm, Will Radford, and Ben Hachey. 2017.
\newblock Learning to generate one-sentence biographies from wikidata.
\newblock \emph{arXiv preprint arXiv:1702.06235}.

\bibitem[{Dai et~al.(2019)Dai, Yang, Yang, Carbonell, Le, and
  Salakhutdinov}]{dai2019transformer}
Zihang Dai, Zhilin Yang, Yiming Yang, Jaime Carbonell, Quoc~V Le, and Ruslan
  Salakhutdinov. 2019.
\newblock Transformer-xl: Attentive language models beyond a fixed-length
  context.
\newblock \emph{arXiv preprint arXiv:1901.02860}.

\bibitem[{De-Arteaga et~al.(2019)De-Arteaga, Romanov, Wallach, Chayes, Borgs,
  Chouldechova, Geyik, Kenthapadi, and Kalai}]{de2019bias}
Maria De-Arteaga, Alexey Romanov, Hanna Wallach, Jennifer Chayes, Christian
  Borgs, Alexandra Chouldechova, Sahin Geyik, Krishnaram Kenthapadi, and
  Adam~Tauman Kalai. 2019.
\newblock Bias in bios: A case study of semantic representation bias in a
  high-stakes setting.
\newblock In \emph{proceedings of the Conference on Fairness, Accountability,
  and Transparency}, pages 120--128.

\bibitem[{Devlin et~al.(2018)Devlin, Chang, Lee, and
  Toutanova}]{devlin2018bert}
Jacob Devlin, Ming-Wei Chang, Kenton Lee, and Kristina Toutanova. 2018.
\newblock Bert: Pre-training of deep bidirectional transformers for language
  understanding.
\newblock \emph{arXiv preprint arXiv:1810.04805}.

\bibitem[{DeYoung et~al.(2019)DeYoung, Jain, Rajani, Lehman, Xiong, Socher, and
  Wallace}]{deyoung2019eraser}
Jay DeYoung, Sarthak Jain, Nazneen~Fatema Rajani, Eric Lehman, Caiming Xiong,
  Richard Socher, and Byron~C Wallace. 2019.
\newblock Eraser: A benchmark to evaluate rationalized nlp models.
\newblock \emph{arXiv preprint arXiv:1911.03429}.

\bibitem[{Dinan et~al.(2019)Dinan, Fan, Williams, Urbanek, Kiela, and
  Weston}]{dinan2019queens}
Emily Dinan, Angela Fan, Adina Williams, Jack Urbanek, Douwe Kiela, and Jason
  Weston. 2019.
\newblock Queens are powerful too: Mitigating gender bias in dialogue
  generation.
\newblock \emph{arXiv preprint arXiv:1911.03842}.

\bibitem[{Dinan et~al.(2020)Dinan, Fan, Wu, Weston, Kiela, and
  Williams}]{dinan2020multi}
Emily Dinan, Angela Fan, Ledell Wu, Jason Weston, Douwe Kiela, and Adina
  Williams. 2020.
\newblock Multi-dimensional gender bias classification.
\newblock \emph{arXiv preprint arXiv:2005.00614}.

\bibitem[{Dinan et~al.(2018)Dinan, Roller, Shuster, Fan, Auli, and
  Weston}]{dinan2018wizard}
Emily Dinan, Stephen Roller, Kurt Shuster, Angela Fan, Michael Auli, and Jason
  Weston. 2018.
\newblock Wizard of wikipedia: Knowledge-powered conversational agents.
\newblock \emph{arXiv preprint arXiv:1811.01241}.

\bibitem[{Du{\v{s}}ek and Kasner(2020)}]{duvsek2020evaluating}
Ond{\v{r}}ej Du{\v{s}}ek and Zden{\v{e}}k Kasner. 2020.
\newblock Evaluating semantic accuracy of data-to-text generation with natural
  language inference.
\newblock \emph{arXiv preprint arXiv:2011.10819}.

\bibitem[{Elazar et~al.(2021)Elazar, Kassner, Ravfogel, Ravichander, Hovy,
  Sch{\"u}tze, and Goldberg}]{elazar2021measuring}
Yanai Elazar, Nora Kassner, Shauli Ravfogel, Abhilasha Ravichander, Eduard
  Hovy, Hinrich Sch{\"u}tze, and Yoav Goldberg. 2021.
\newblock Measuring and improving consistency in pretrained language models.
\newblock \emph{Transactions of the Association for Computational Linguistics},
  9:1012--1031.

\bibitem[{Fan et~al.(2019{\natexlab{a}})Fan, Gardent, Braud, and
  Bordes}]{fan2019using}
Angela Fan, Claire Gardent, Chlo{\'e} Braud, and Antoine Bordes.
  2019{\natexlab{a}}.
\newblock Using local knowledge graph construction to scale seq2seq models to
  multi-document inputs.
\newblock In \emph{2019 Conference on Empirical Methods in Natural Language
  Processing and 9th International Joint Conference on Natural Language
  Processing}.

\bibitem[{Fan et~al.(2019{\natexlab{b}})Fan, Grave, and
  Joulin}]{fan2019reducing}
Angela Fan, Edouard Grave, and Armand Joulin. 2019{\natexlab{b}}.
\newblock Reducing transformer depth on demand with structured dropout.
\newblock \emph{arXiv preprint arXiv:1909.11556}.

\bibitem[{Fan et~al.(2019{\natexlab{c}})Fan, Jernite, Perez, Grangier, Weston,
  and Auli}]{fan2019eli5}
Angela Fan, Yacine Jernite, Ethan Perez, David Grangier, Jason Weston, and
  Michael Auli. 2019{\natexlab{c}}.
\newblock Eli5: Long form question answering.
\newblock \emph{arXiv preprint arXiv:1907.09190}.

\bibitem[{Gardent et~al.(2017)Gardent, Shimorina, Narayan, and
  Perez-Beltrachini}]{gardent-etal-2017-webnlg}
Claire Gardent, Anastasia Shimorina, Shashi Narayan, and Laura
  Perez-Beltrachini. 2017.
\newblock \href {https://doi.org/10.18653/v1/W17-3518} {The {W}eb{NLG}
  challenge: Generating text from {RDF} data}.
\newblock In \emph{Proceedings of the 10th International Conference on Natural
  Language Generation}, pages 124--133, Santiago de Compostela, Spain.
  Association for Computational Linguistics.

\bibitem[{Gonen and Goldberg(2019)}]{gonen2019lipstick}
Hila Gonen and Yoav Goldberg. 2019.
\newblock Lipstick on a pig: Debiasing methods cover up systematic gender
  biases in word embeddings but do not remove them.
\newblock \emph{arXiv preprint arXiv:1903.03862}.

\bibitem[{Gonzalez et~al.(2020)Gonzalez, Bansal, Fan, Jia, Mehdad, and
  Iyer}]{gonzalez2020human}
Ana~Valeria Gonzalez, Gagan Bansal, Angela Fan, Robin Jia, Yashar Mehdad, and
  Srinivasan Iyer. 2020.
\newblock Human evaluation of spoken vs. visual explanations for open-domain
  qa.
\newblock \emph{arXiv preprint arXiv:2012.15075}.

\bibitem[{Goodrich et~al.(2019)Goodrich, Rao, Liu, and
  Saleh}]{goodrich2019assessing}
Ben Goodrich, Vinay Rao, Peter~J Liu, and Mohammad Saleh. 2019.
\newblock Assessing the factual accuracy of generated text.
\newblock In \emph{Proceedings of the 25th ACM SIGKDD International Conference
  on Knowledge Discovery \& Data Mining}, pages 166--175.

\bibitem[{Graells-Garrido et~al.(2015)Graells-Garrido, Lalmas, and
  Menczer}]{graells2015first}
Eduardo Graells-Garrido, Mounia Lalmas, and Filippo Menczer. 2015.
\newblock First women, second sex: Gender bias in wikipedia.
\newblock In \emph{Proceedings of the 26th ACM Conference on Hypertext \&
  Social Media}, pages 165--174.

\bibitem[{Guu et~al.(2018)Guu, Hashimoto, Oren, and Liang}]{guu2018generating}
Kelvin Guu, Tatsunori~B Hashimoto, Yonatan Oren, and Percy Liang. 2018.
\newblock Generating sentences by editing prototypes.
\newblock \emph{Transactions of the Association for Computational Linguistics},
  6:437--450.

\bibitem[{Guu et~al.(2020)Guu, Lee, Tung, Pasupat, and Chang}]{guu2020realm}
Kelvin Guu, Kenton Lee, Zora Tung, Panupong Pasupat, and Ming-Wei Chang. 2020.
\newblock Realm: Retrieval-augmented language model pre-training.
\newblock \emph{arXiv preprint arXiv:2002.08909}.

\bibitem[{Hase et~al.(2020)Hase, Zhang, Xie, and Bansal}]{hase2020leakage}
Peter Hase, Shiyue Zhang, Harry Xie, and Mohit Bansal. 2020.
\newblock Leakage-adjusted simulatability: Can models generate non-trivial
  explanations of their behavior in natural language?
\newblock \emph{arXiv preprint arXiv:2010.04119}.

\bibitem[{Hinnosaar(2019)}]{hinnosaar2019gender}
Marit Hinnosaar. 2019.
\newblock Gender inequality in new media: Evidence from wikipedia.
\newblock \emph{Journal of Economic Behavior \& Organization}, 163:262--276.

\bibitem[{Howcroft et~al.(2020)Howcroft, Belz, Clinciu, Gkatzia, Hasan,
  Mahamood, Mille, van Miltenburg, Santhanam, and Rieser}]{howcroft2020twenty}
David~M Howcroft, Anja Belz, Miruna-Adriana Clinciu, Dimitra Gkatzia, Sadid~A
  Hasan, Saad Mahamood, Simon Mille, Emiel van Miltenburg, Sashank Santhanam,
  and Verena Rieser. 2020.
\newblock Twenty years of confusion in human evaluation: Nlg needs evaluation
  sheets and standardised definitions.
\newblock In \emph{Proceedings of the 13th International Conference on Natural
  Language Generation}, pages 169--182.

\bibitem[{Iglesias(2020)}]{iglesias2020preparing}
Encina~Calvo Iglesias. 2020.
\newblock Preparing biographies of stem women in the wikipedia format, a
  teaching experience.
\newblock \emph{IEEE Revista Iberoamericana de Tecnologias del Aprendizaje},
  15(3):211--214.

\bibitem[{Izacard and Grave(2020)}]{izacard2020leveraging}
Gautier Izacard and Edouard Grave. 2020.
\newblock Leveraging passage retrieval with generative models for open domain
  question answering.
\newblock \emph{arXiv preprint arXiv:2007.01282}.

\bibitem[{Jin et~al.(2020)Jin, Guo, Qiu, and Zhang}]{jin2020genwiki}
Zhijing Jin, Qipeng Guo, Xipeng Qiu, and Zheng Zhang. 2020.
\newblock Genwiki: A dataset of 1.3 million content-sharing text and graphs for
  unsupervised graph-to-text generation.
\newblock In \emph{Proceedings of the 28th International Conference on
  Computational Linguistics}, pages 2398--2409.

\bibitem[{Kaffee et~al.(2018{\natexlab{a}})Kaffee, Elsahar, Vougiouklis,
  Gravier, Laforest, Hare, and Simperl}]{kaffee2018learning}
Lucie-Aim{\'e}e Kaffee, Hady Elsahar, Pavlos Vougiouklis, Christophe Gravier,
  Fr{\'e}d{\'e}rique Laforest, Jonathon Hare, and Elena Simperl.
  2018{\natexlab{a}}.
\newblock Learning to generate wikipedia summaries for underserved languages
  from wikidata.
\newblock \emph{arXiv preprint arXiv:1803.07116}.

\bibitem[{Kaffee et~al.(2018{\natexlab{b}})Kaffee, Elsahar, Vougiouklis,
  Gravier, Laforest, Hare, and Simperl}]{kaffee2018mind}
Lucie-Aim{\'e}e Kaffee, Hady Elsahar, Pavlos Vougiouklis, Christophe Gravier,
  Fr{\'e}d{\'e}rique Laforest, Jonathon Hare, and Elena Simperl.
  2018{\natexlab{b}}.
\newblock Mind the (language) gap: Generation of multilingual wikipedia
  summaries from wikidata for articleplaceholders.
\newblock In \emph{European Semantic Web Conference}, pages 319--334. Springer.

\bibitem[{Kaffee et~al.(2020)Kaffee, Vougiouklis, and
  Simperl}]{kaffee2020using}
Lucie-Aim{\'e}e Kaffee, Pavlos Vougiouklis, and Elena Simperl. 2020.
\newblock Using natural language generation to bootstrap missing wikipedia
  articles: A human-centric perspective.
\newblock \emph{Semantic Web Journal}.

\bibitem[{Karpukhin et~al.(2020)Karpukhin, Oguz, Min, Lewis, Wu, Edunov, Chen,
  and Yih}]{karpukhin-etal-2020-dense}
Vladimir Karpukhin, Barlas Oguz, Sewon Min, Patrick Lewis, Ledell Wu, Sergey
  Edunov, Danqi Chen, and Wen-tau Yih. 2020.
\newblock \href {https://doi.org/10.18653/v1/2020.emnlp-main.550} {Dense
  passage retrieval for open-domain question answering}.
\newblock In \emph{Proceedings of the 2020 Conference on Empirical Methods in
  Natural Language Processing (EMNLP)}, pages 6769--6781, Online. Association
  for Computational Linguistics.

\bibitem[{Komeili et~al.(2021)Komeili, Shuster, and
  Weston}]{komeili2021internet}
Mojtaba Komeili, Kurt Shuster, and Jason Weston. 2021.
\newblock Internet-augmented dialogue generation.
\newblock \emph{arXiv preprint arXiv:2107.07566}.

\bibitem[{Kumar and Talukdar(2020)}]{kumar2020nile}
Sawan Kumar and Partha Talukdar. 2020.
\newblock Nile: Natural language inference with faithful natural language
  explanations.
\newblock \emph{arXiv preprint arXiv:2005.12116}.

\bibitem[{Kwiatkowski et~al.(2019)Kwiatkowski, Palomaki, Redfield, Collins,
  Parikh, Alberti, Epstein, Polosukhin, Devlin, Lee
  et~al.}]{kwiatkowski2019natural}
Tom Kwiatkowski, Jennimaria Palomaki, Olivia Redfield, Michael Collins, Ankur
  Parikh, Chris Alberti, Danielle Epstein, Illia Polosukhin, Jacob Devlin,
  Kenton Lee, et~al. 2019.
\newblock Natural questions: a benchmark for question answering research.
\newblock \emph{Transactions of the Association for Computational Linguistics},
  7:453--466.

\bibitem[{Lakhotia et~al.(2020)Lakhotia, Paranjape, Ghoshal, Yih, Mehdad, and
  Iyer}]{lakhotia2020fid}
Kushal Lakhotia, Bhargavi Paranjape, Asish Ghoshal, Wen-tau Yih, Yashar Mehdad,
  and Srinivasan Iyer. 2020.
\newblock Fid-ex: Improving sequence-to-sequence models for extractive
  rationale generation.
\newblock \emph{arXiv preprint arXiv:2012.15482}.

\bibitem[{Lamm et~al.(2020)Lamm, Palomaki, Alberti, Andor, Choi, Soares, and
  Collins}]{lamm2020qed}
Matthew Lamm, Jennimaria Palomaki, Chris Alberti, Daniel Andor, Eunsol Choi,
  Livio~Baldini Soares, and Michael Collins. 2020.
\newblock Qed: A framework and dataset for explanations in question answering.
\newblock \emph{arXiv preprint arXiv:2009.06354}.

\bibitem[{Langrock and Gonz{\'a}lez-Bail{\'o}n(2020)}]{langrock2020gender}
Isabelle Langrock and Sandra Gonz{\'a}lez-Bail{\'o}n. 2020.
\newblock The gender divide in wikipedia: A computational approach to assessing
  the impact of two feminist interventions.
\newblock \emph{Available at SSRN}.

\bibitem[{Latcinnik and Berant(2020)}]{latcinnik2020explaining}
Veronica Latcinnik and Jonathan Berant. 2020.
\newblock Explaining question answering models through text generation.
\newblock \emph{arXiv preprint arXiv:2004.05569}.

\bibitem[{Lebret et~al.(2016)Lebret, Grangier, and Auli}]{lebret2016neural}
R{\'e}mi Lebret, David Grangier, and Michael Auli. 2016.
\newblock Neural text generation from structured data with application to the
  biography domain.
\newblock \emph{arXiv preprint arXiv:1603.07771}.

\bibitem[{Lee et~al.(2019)Lee, Madotto, and Fung}]{lee2019exploring}
Nayeon Lee, Andrea Madotto, and Pascale Fung. 2019.
\newblock Exploring social bias in chatbots using stereotype knowledge.
\newblock In \emph{Proceedings of the 2019 Workshop on Widening NLP}, pages
  177--180.

\bibitem[{Lewis et~al.(2019)Lewis, Liu, Goyal, Ghazvininejad, Mohamed, Levy,
  Stoyanov, and Zettlemoyer}]{lewis2019bart}
Mike Lewis, Yinhan Liu, Naman Goyal, Marjan Ghazvininejad, Abdelrahman Mohamed,
  Omer Levy, Ves Stoyanov, and Luke Zettlemoyer. 2019.
\newblock Bart: Denoising sequence-to-sequence pre-training for natural
  language generation, translation, and comprehension.
\newblock \emph{arXiv preprint arXiv:1910.13461}.

\bibitem[{Lewis et~al.(2020)Lewis, Perez, Piktus, Petroni, Karpukhin, Goyal,
  K{\"u}ttler, Lewis, Yih, Rockt{\"a}schel et~al.}]{lewis2020retrieval}
Patrick Lewis, Ethan Perez, Aleksandara Piktus, Fabio Petroni, Vladimir
  Karpukhin, Naman Goyal, Heinrich K{\"u}ttler, Mike Lewis, Wen-tau Yih, Tim
  Rockt{\"a}schel, et~al. 2020.
\newblock Retrieval-augmented generation for knowledge-intensive nlp tasks.
\newblock \emph{arXiv preprint arXiv:2005.11401}.

\bibitem[{Lin(2004)}]{lin2004rouge}
Chin-Yew Lin. 2004.
\newblock Rouge: A package for automatic evaluation of summaries.
\newblock In \emph{Text summarization branches out}, pages 74--81.

\bibitem[{Liu et~al.(2020)Liu, Wang, Wang, Liu, Liu, and
  Tang}]{liu2020mitigating}
Haochen Liu, Wentao Wang, Yiqi Wang, Hui Liu, Zitao Liu, and Jiliang Tang.
  2020.
\newblock Mitigating gender bias for neural dialogue generation with
  adversarial learning.
\newblock \emph{arXiv preprint arXiv:2009.13028}.

\bibitem[{Liu et~al.(2018)Liu, Saleh, Pot, Goodrich, Sepassi, Kaiser, and
  Shazeer}]{liu2018generating}
Peter~J Liu, Mohammad Saleh, Etienne Pot, Ben Goodrich, Ryan Sepassi, Lukasz
  Kaiser, and Noam Shazeer. 2018.
\newblock Generating wikipedia by summarizing long sequences.
\newblock \emph{arXiv preprint arXiv:1801.10198}.

\bibitem[{Liu et~al.(2021)Liu, Zhang, Brockett, Mao, Sui, Chen, and
  Dolan}]{liu2021token}
Tianyu Liu, Yizhe Zhang, Chris Brockett, Yi~Mao, Zhifang Sui, Weizhu Chen, and
  Bill Dolan. 2021.
\newblock A token-level reference-free hallucination detection benchmark for
  free-form text generation.
\newblock \emph{arXiv preprint arXiv:2104.08704}.

\bibitem[{Liu et~al.(2010)Liu, Nie, Yu, and Wen}]{liu2010biosnowball}
Xiaojiang Liu, Zaiqing Nie, Nenghai Yu, and Ji-Rong Wen. 2010.
\newblock Biosnowball: automated population of wikis.
\newblock In \emph{Proceedings of the 16th ACM SIGKDD international conference
  on Knowledge discovery and data mining}, pages 969--978.

\bibitem[{Liu et~al.(2019)Liu, Ott, Goyal, Du, Joshi, Chen, Levy, Lewis,
  Zettlemoyer, and Stoyanov}]{liu2019roberta}
Yinhan Liu, Myle Ott, Naman Goyal, Jingfei Du, Mandar Joshi, Danqi Chen, Omer
  Levy, Mike Lewis, Luke Zettlemoyer, and Veselin Stoyanov. 2019.
\newblock Roberta: A robustly optimized bert pretraining approach.
\newblock \emph{arXiv preprint arXiv:1907.11692}.

\bibitem[{Luo et~al.(2018)Luo, Adams, and Brueckner}]{luo2018ladies}
Wei Luo, Julia Adams, and Hannah Brueckner. 2018.
\newblock The ladies vanish?: American sociology and the genealogy of its
  missing women on wikipedia.
\newblock \emph{Comparative Sociology}, 17(5):519--556.

\bibitem[{Maynez et~al.(2020)Maynez, Narayan, Bohnet, and
  McDonald}]{maynez2020faithfulness}
Joshua Maynez, Shashi Narayan, Bernd Bohnet, and Ryan McDonald. 2020.
\newblock On faithfulness and factuality in abstractive summarization.
\newblock \emph{arXiv preprint arXiv:2005.00661}.

\bibitem[{Moghe et~al.(2018)Moghe, Arora, Banerjee, and
  Khapra}]{moghe2018towards}
Nikita Moghe, Siddhartha Arora, Suman Banerjee, and Mitesh~M Khapra. 2018.
\newblock Towards exploiting background knowledge for building conversation
  systems.
\newblock \emph{arXiv preprint arXiv:1809.08205}.

\bibitem[{Narang et~al.(2020)Narang, Raffel, Lee, Roberts, Fiedel, and
  Malkan}]{narang2020wt5}
Sharan Narang, Colin Raffel, Katherine Lee, Adam Roberts, Noah Fiedel, and
  Karishma Malkan. 2020.
\newblock Wt5?! training text-to-text models to explain their predictions.
\newblock \emph{arXiv preprint arXiv:2004.14546}.

\bibitem[{Nie et~al.(2019)Nie, Yao, Wang, Pan, and Lin}]{nie2019simple}
Feng Nie, Jin-Ge Yao, Jinpeng Wang, Rong Pan, and Chin-Yew Lin. 2019.
\newblock A simple recipe towards reducing hallucination in neural surface
  realisation.
\newblock In \emph{Proceedings of the 57th Annual Meeting of the Association
  for Computational Linguistics}, pages 2673--2679.

\bibitem[{Parikh et~al.(2020)Parikh, Wang, Gehrmann, Faruqui, Dhingra, Yang,
  and Das}]{parikh2020totto}
Ankur~P Parikh, Xuezhi Wang, Sebastian Gehrmann, Manaal Faruqui, Bhuwan
  Dhingra, Diyi Yang, and Dipanjan Das. 2020.
\newblock Totto: A controlled table-to-text generation dataset.
\newblock \emph{arXiv preprint arXiv:2004.14373}.

\bibitem[{Park et~al.(2018)Park, Shin, and Fung}]{park2018reducing}
Ji~Ho Park, Jamin Shin, and Pascale Fung. 2018.
\newblock Reducing gender bias in abusive language detection.
\newblock \emph{arXiv preprint arXiv:1808.07231}.

\bibitem[{Peshterliev et~al.(2021)Peshterliev, Oguz, Chatterjee, Inan, and
  Bhardwaj}]{peshterliev2021conversational}
Stan Peshterliev, Barlas Oguz, Debojeet Chatterjee, Hakan Inan, and Vikas
  Bhardwaj. 2021.
\newblock Conversational answer generation and factuality for reading
  comprehension question-answering.
\newblock \emph{arXiv preprint arXiv:2103.06500}.

\bibitem[{Piktus et~al.(2021)Piktus, Petroni, Karpukhin, Okhonko, Broscheit,
  Izacard, Lewis, O{\u{g}}uz, Grave, Yih et~al.}]{piktus2021web}
Aleksandra Piktus, Fabio Petroni, Vladimir Karpukhin, Dmytro Okhonko, Samuel
  Broscheit, Gautier Izacard, Patrick Lewis, Barlas O{\u{g}}uz, Edouard Grave,
  Wen-tau Yih, et~al. 2021.
\newblock The web is your oyster--knowledge-intensive nlp against a very large
  web corpus.
\newblock \emph{arXiv preprint arXiv:2112.09924}.

\bibitem[{Puduppully et~al.(2019)Puduppully, Dong, and
  Lapata}]{puduppully2019data}
Ratish Puduppully, Li~Dong, and Mirella Lapata. 2019.
\newblock Data-to-text generation with content selection and planning.
\newblock In \emph{Proceedings of the AAAI conference on artificial
  intelligence}, volume~33, pages 6908--6915.

\bibitem[{Radford et~al.(2019)Radford, Wu, Child, Luan, Amodei, Sutskever
  et~al.}]{radford2019language}
Alec Radford, Jeffrey Wu, Rewon Child, David Luan, Dario Amodei, Ilya
  Sutskever, et~al. 2019.
\newblock Language models are unsupervised multitask learners.
\newblock \emph{OpenAI blog}, 1(8):9.

\bibitem[{Sauper and Barzilay(2009)}]{sauper2009automatically}
Christina~Joan Sauper and Regina Barzilay. 2009.
\newblock Automatically generating wikipedia articles: A structure-aware
  approach.
\newblock Association for Computational Linguistics.

\bibitem[{Schmahl et~al.(2020)Schmahl, Viering, Makrodimitris, Jahfari, Tax,
  and Loog}]{schmahl2020wikipedia}
Katja~Geertruida Schmahl, Tom~Julian Viering, Stavros Makrodimitris,
  Arman~Naseri Jahfari, David Tax, and Marco Loog. 2020.
\newblock Is wikipedia succeeding in reducing gender bias? assessing changes in
  gender bias in wikipedia using word embeddings.
\newblock In \emph{Proceedings of the Fourth Workshop on Natural Language
  Processing and Computational Social Science}, pages 94--103.

\bibitem[{Seo et~al.(2019)Seo, Lee, Kwiatkowski, Parikh, Farhadi, and
  Hajishirzi}]{seo2019real}
Minjoon Seo, Jinhyuk Lee, Tom Kwiatkowski, Ankur~P Parikh, Ali Farhadi, and
  Hannaneh Hajishirzi. 2019.
\newblock Real-time open-domain question answering with dense-sparse phrase
  index.
\newblock \emph{arXiv preprint arXiv:1906.05807}.

\bibitem[{Sha et~al.(2018)Sha, Mou, Liu, Poupart, Li, Chang, and
  Sui}]{sha2018order}
Lei Sha, Lili Mou, Tianyu Liu, Pascal Poupart, Sujian Li, Baobao Chang, and
  Zhifang Sui. 2018.
\newblock Order-planning neural text generation from structured data.
\newblock In \emph{Thirty-Second AAAI Conference on Artificial Intelligence}.

\bibitem[{Shuster et~al.(2022)Shuster, Komeili, Adolphs, Roller, Szlam, and
  Weston}]{shuster2022language}
Kurt Shuster, Mojtaba Komeili, Leonard Adolphs, Stephen Roller, Arthur Szlam,
  and Jason Weston. 2022.
\newblock Language models that seek for knowledge: Modular search \& generation
  for dialogue and prompt completion.
\newblock \emph{arXiv preprint arXiv:2203.13224}.

\bibitem[{Shuster et~al.(2021)Shuster, Poff, Chen, Kiela, and
  Weston}]{shuster2021retrieval}
Kurt Shuster, Spencer Poff, Moya Chen, Douwe Kiela, and Jason Weston. 2021.
\newblock Retrieval augmentation reduces hallucination in conversation.
\newblock \emph{arXiv preprint arXiv:2104.07567}.

\bibitem[{Stanovsky et~al.(2019)Stanovsky, Smith, and
  Zettlemoyer}]{stanovsky2019evaluating}
Gabriel Stanovsky, Noah~A Smith, and Luke Zettlemoyer. 2019.
\newblock Evaluating gender bias in machine translation.
\newblock \emph{arXiv preprint arXiv:1906.00591}.

\bibitem[{Stratigakos(2016)}]{stratigakos2016unforgetting}
Despina Stratigakos. 2016.
\newblock Unforgetting women architects: From the pritzker to wikipedia.
\newblock \emph{Places Journal}.

\bibitem[{Thomson and Reiter(2020)}]{thomson2020gold}
Craig Thomson and Ehud Reiter. 2020.
\newblock A gold standard methodology for evaluating accuracy in data-to-text
  systems.
\newblock \emph{arXiv preprint arXiv:2011.03992}.

\bibitem[{Thorne et~al.(2018)Thorne, Vlachos, Christodoulopoulos, and
  Mittal}]{thorne2018fever}
James Thorne, Andreas Vlachos, Christos Christodoulopoulos, and Arpit Mittal.
  2018.
\newblock Fever: a large-scale dataset for fact extraction and verification.
\newblock \emph{arXiv preprint arXiv:1803.05355}.

\bibitem[{Tian et~al.(2019)Tian, Narayan, Sellam, and
  Parikh}]{tian2019sticking}
Ran Tian, Shashi Narayan, Thibault Sellam, and Ankur~P Parikh. 2019.
\newblock Sticking to the facts: Confident decoding for faithful data-to-text
  generation.
\newblock \emph{arXiv preprint arXiv:1910.08684}.

\bibitem[{Vougiouklis et~al.(2018)Vougiouklis, Elsahar, Kaffee, Gravier,
  Laforest, Hare, and Simperl}]{vougiouklis2018neural}
Pavlos Vougiouklis, Hady Elsahar, Lucie-Aim{\'e}e Kaffee, Christophe Gravier,
  Fr{\'e}d{\'e}rique Laforest, Jonathon Hare, and Elena Simperl. 2018.
\newblock Neural wikipedian: Generating textual summaries from knowledge base
  triples.
\newblock \emph{Journal of Web Semantics}, 52:1--15.

\bibitem[{Wang et~al.(2020)Wang, Wang, An, Yu, and Chen}]{wang2020towards}
Zhenyi Wang, Xiaoyang Wang, Bang An, Dong Yu, and Changyou Chen. 2020.
\newblock Towards faithful neural table-to-text generation with
  content-matching constraints.
\newblock \emph{arXiv preprint arXiv:2005.00969}.

\bibitem[{Worku et~al.(2020)Worku, Bipat, McDonald, and
  Zachry}]{worku2020exploring}
Zena Worku, Taryn Bipat, David~W McDonald, and Mark Zachry. 2020.
\newblock Exploring systematic bias through article deletions on wikipedia from
  a behavioral perspective.
\newblock In \emph{Proceedings of the 16th International Symposium on Open
  Collaboration}, pages 1--22.

\bibitem[{Wu et~al.(2019)Wu, Petroni, Josifoski, Riedel, and
  Zettlemoyer}]{wu2019scalable}
Ledell Wu, Fabio Petroni, Martin Josifoski, Sebastian Riedel, and Luke
  Zettlemoyer. 2019.
\newblock Scalable zero-shot entity linking with dense entity retrieval.
\newblock \emph{arXiv preprint arXiv:1911.03814}.

\bibitem[{Yeo and Chen(2020)}]{yeo2020defining}
Catherine Yeo and Alyssa Chen. 2020.
\newblock Defining and evaluating fair natural language generation.
\newblock \emph{arXiv preprint arXiv:2008.01548}.

\bibitem[{Young et~al.(2020)Young, Wigdor, and Kane}]{young2020gender}
Amber~G Young, Ariel~D Wigdor, and Gerald~C Kane. 2020.
\newblock The gender bias tug-of-war in a co-creation community: Core-periphery
  tension on wikipedia.
\newblock \emph{Journal of Management Information Systems}, 37(4):1047--1072.

\bibitem[{Zhao et~al.(2018)Zhao, Wang, Yatskar, Ordonez, and
  Chang}]{zhao2018gender}
Jieyu Zhao, Tianlu Wang, Mark Yatskar, Vicente Ordonez, and Kai-Wei Chang.
  2018.
\newblock Gender bias in coreference resolution: Evaluation and debiasing
  methods.
\newblock \emph{arXiv preprint arXiv:1804.06876}.

\bibitem[{Zhou et~al.(2020)Zhou, Neubig, Gu, Diab, Guzman, Zettlemoyer, and
  Ghazvininejad}]{zhou2020detecting}
Chunting Zhou, Graham Neubig, Jiatao Gu, Mona Diab, Paco Guzman, Luke
  Zettlemoyer, and Marjan Ghazvininejad. 2020.
\newblock Detecting hallucinated content in conditional neural sequence
  generation.
\newblock \emph{arXiv preprint arXiv:2011.02593}.

\end{thebibliography}
\bibliographystyle{acl_natbib}

\appendix
\clearpage 
\newpage 
\section{Appendix}

\begin{figure*}[t]
  \small
  \rule{\linewidth}{1pt}
     \textbf{toplevel} phoebe legere is an american singer - songwriter, painter, actress, and musician. she is a member of the acadian - cajun renaissance, a group of artists who combine elements of americana, cajuns, blues, and jazz. she has released 17 cds of original and traditional music, and her latest album is heart of love, and on the charts on americana radio. she appears on hbo 's documentary it' s me hilary. her original song "hip hop frog ", a song about the environment, was licensed by hbo. leger 's paintings and drawings are in the collections of the metropolitan museum of art, the museum of fine arts, boston, and the metropolitan opera. [1,3,4,8,11]
    \rule{\linewidth}{1pt}
     \textbf{toplevel} joan paton (née paton) (c. 1883 – 1962) was a british ornithologist. she was the first woman to be elected a fellow of the linnean society of london and the royal society of edinburgh. she is best known for her work in the field of ornithology, particularly her work on the birds of wales and scandinavia, and for her contributions to the study of birds of the north of england and scotland. [2]
    \\
    \textbf{=early life=} paton was born in london, england. she was educated at the university of cambridge, where she graduated with a bachelor of arts (ba) degree in zoology and a master of science (msc) degree. she went on to earn a doctor of philosophy (phd) degree from the london school of hygiene and tropical medicine (lse) and a postgraduate diploma in tropical medicine and hygiene from the royal college of physicians and surgeons of london (rcpsl). [2,5]  
    \\
     \textbf{=career=} paton began her career as an ornithologist at the royal botanic gardens, kew, where she was a member of the ornithological society of london. she was elected a fellow of the british ornithologists' union (f. a. e. u.) in 1954. she served as the society 's vice - president from 1958 to 1960. she became a fellow in 1962 and was elected to the royal society of edinburgh in 1964. she also served on the council of the society for the protection of birds of great britain and ireland. paton was elected an honorary fellow of st john 's college, cambridge in 1966. she retired from the society in 1972. she died in london in 1984. [1,2]
    \rule{\linewidth}{1pt}
     \textbf{toplevel} ashley mckenzie is a canadian film director, screenwriter and producer. she is the winner of the stella artois jay scott prize for emerging talent at the 2016 toronto international film festival. her first feature film, werewolf, premiered at the toronto film festival in 2016. she has also directed short films for the national film board of canada and the canadian screen actors guild. she was born in montreal, quebec, canada, and grew up in ottawa, ontario. [1,3,11,13,14]
    \\
     \textbf{=personal life=} mckenzie was born in london, england. she is the daughter of alexander mckenzie, who was a member of the british rock band the beatles. she has a younger sister, jessica, who is also a singer. she was educated at st mary 's college, oxford, where she graduated with a bachelor of arts degree in english literature. she also studied at the university of london. she married fellow x factor contestant andrew davies in september 2006. they have two children, a son and a daughter. [3,4,7,8,10,11]                                                    
    \\
     \textbf{=career=} mckenzie was a contestant on the third series of the x - factor in 2006. she was eliminated in the first week of the competition. in 2007, mckenzie released her debut single "don 't pretend you hadn' t, now..." which peaked at no .160; 2 on the uk singles chart. she also released a second single ," i 'm not afraid ", in 2008. in 2009, she released her third single ," don' t pretend you haven 't, now ". in 2010, she was a judge on the x factor uk. [2]
  \rule{\linewidth}{1pt}
  \caption{\textbf{Random Examples of Generated Articles.} Note that \textit{toplevel} is an augmented special tag to indicate the start of the article and \textit{=} surrounds section headings on Wikipedia. Text in brackets indicates the cited references.}
  \label{fig:examples_wikipedia_generation}
\end{figure*} 

\subsection{Model and Training Details}

We use the BART-Large model as open sourced by~\citet{lewis2019bart}. We train with learning rate $3e-05$ and a polynomial decay learning rate schedule, warming up for 500 updates, and end training after 50,000 updates. We train with dropout and attention dropout $0.1$, label smoothing $0.1$, and $0.01$ weight decay. Our final model trains on 8 GPUs for three days. For experimentation, we train on 4 GPUs for 12 hours, which is about the time required for convergence. 

\subsection{Human Evaluation Details}

Our evaluation is conducted on the Amazon Mechanical Turk platform. We pay evaluators approximately fifteen dollars an hour. Each section is evaluated independently, and evaluation tasks are not batched. The generated section and reference section are displayed side by side, segmented into separate sentences. To ease the challenge of human evaluation, we evaluate sentence by sentence. This is displayed by highlighting sentences independently, to reduce information overload. 

\subsection{Additional Examples} 

We present several examples of full generated articles in Figure~\ref{fig:examples_wikipedia_generation}. 

\subsection{Amount of Information Used from Retrieved Documents} 

Sequence-to-sequence models for text generation are able to utilize retrieval to augment generation, widely used in tasks such as question answering. 
Compared to these tasks, where the information to e.g. compose a written answer to a question is contained in a very specific paragraph, writing Wikipedia articles is much more freeform.
For example, Wikipedia articles usually are written by human editors who have looked at a large amount of source material and paraphrased it, and articles are edited by many people over time. 
Thus, we find that it is difficult to directly retrieve a perfect provenance document that part of the Wikipedia article could be copy-pasted from.

We analyze how the model utilizes the retrieved information, and we find three main cases. In the first case, a small number of the web search documents are very useful (for example, biographical information about the person already on the web, such as on \texttt{biography.com}). In this case, the model utilizes this information very heavily, and often only retrieves content from this small number of documents.
In the second case, there are a number of partially relevant documents, and web searches on the different predicted section headings change the web search results. Thus, models retrieve small amounts of information from multiple different sources.
Finally, the third case is discussed in Section 7.2, and is potentially the most challenging to resolve: the situation where little information about the biographical subject is present on the web. 

These three scenarios arise for all biographical articles, but differ in prevalence between different categories of people. For example, certain occupations more naturally come with some quantity of information available online compared to others. An example is Olympic athletes --- at that level of notability, usually their athletic career is chronicled more by the media, thus making a larger quantity of evidence on the web available. Another example can extend to scientists, where we observed that scientists in the United States tend to have personal websites that collate a lot of information, compared to scientists in other locations. 

\end{document}